\documentclass[letterpaper]{article} 
\usepackage{aaai25}  
\usepackage{times}  
\usepackage{helvet}  
\usepackage{courier}  
\usepackage[hyphens]{url}  
\usepackage{graphicx} 
\usepackage{natbib}  
\usepackage{caption} 
\usepackage{algorithm}
\usepackage{algorithmic}
\usepackage{amssymb}
\usepackage{amsmath} 
\usepackage{booktabs}
\usepackage{multirow}

\nocopyright

\usepackage{newfloat}
\usepackage{listings}
\DeclareCaptionStyle{ruled}{labelfont=normalfont,labelsep=colon,strut=off} 
\floatstyle{ruled}
\newfloat{listing}{tb}{lst}{}
\floatname{listing}{Listing}
\title{PD-APE: A Parallel Decoding Framework\\with Adaptive Position Encoding for 3D Visual Grounding}
\author {
    Chenshu Hou\textsuperscript{\rm 1},
    Liang Peng\textsuperscript{\rm 4},
    Xiaopei Wu\textsuperscript{\rm 3},
    Xiaofei He\textsuperscript{\rm 3,4},
    Wenxiao Wang\textsuperscript{\rm 2},
}
\affiliations {
    \textsuperscript{\rm 1}Polytechnic Institute of Zhejiang University\\
    \textsuperscript{\rm 2}School of Software Technology, Zhejiang University\\
    \textsuperscript{\rm 3}State Key Lab of CAD\&CG, Zhejiang University\\
    \textsuperscript{\rm 4}FABU Inc.
}

\usepackage{bibentry}

\begin{document}

\maketitle

\begin{abstract}
3D visual grounding aims to identify objects in 3D point cloud scenes that match specific natural language descriptions. This requires the model to not only focus on the target object itself but also to consider the surrounding environment to determine whether the descriptions are met. Most previous works attempt to accomplish both tasks within the same module, which can easily lead to a distraction of attention. To this end, we propose PD-APE, a dual-branch decoding framework that separately decodes target object attributes and surrounding layouts. Specifically, in the target object branch, the decoder processes text tokens that describe features of the target object (e.g., category and color), guiding the queries to pay attention to the target object itself. In the surrounding branch, the queries align with other text tokens that carry surrounding environment information, making the attention maps accurately capture the layout described in the text. Benefiting from the proposed dual-branch design, the queries are allowed to focus on points relevant to each branch's specific objective. Moreover, we design an adaptive position encoding method for each branch respectively. In the target object branch, the position encoding relies on the relative positions between seed points and predicted 3D boxes. In the surrounding branch, the attention map is additionally guided by the confidence between visual and text features, enabling the queries to focus on points that have valuable layout information. Extensive experiments demonstrate that we surpass the state-of-the-art on two widely adopted 3D visual grounding datasets, ScanRefer and Nr3D.

\end{abstract}

\section{Introduction}

Multi-modal learning is the key to enhancing artificial intelligence's comprehension and awareness of the real world, where the most fundamental and important modalities are visual and language.
Visual Grounding (VG), an emerging vision-language task, requires a model to identify and locate target objects in scenes based on natural language descriptions. 
Although Visual Grounding for 2D images has achieved remarkable results, the sparse and complex structure of 3D point clouds, combined with the linguistic diversity introduced by 3D spatial contexts, presents significant challenges for 3D Visual Grounding.

With significant advances in single-modal 3D object detection, 
the main objective of the 3D Visual Grounding task is to address the issue of ``correct classification but incorrect localization'', which requires the model to better understand the spatial layout of the 3D scene.
Some methods \cite{wang2023drpnet,shi2024aware} attempt to process visual features to better perceive the spatial layouts among detected objects in the scene.
Others \cite{luo20223d,wu2023eda,guo2023viewrefer} focus on processing the natural language descriptions, using techniques such as text decoupling and referential order awareness to achieve positional and semantic cross-modal alignment.

\begin{figure}[!t]
\centering
\includegraphics[width=0.46 \textwidth]{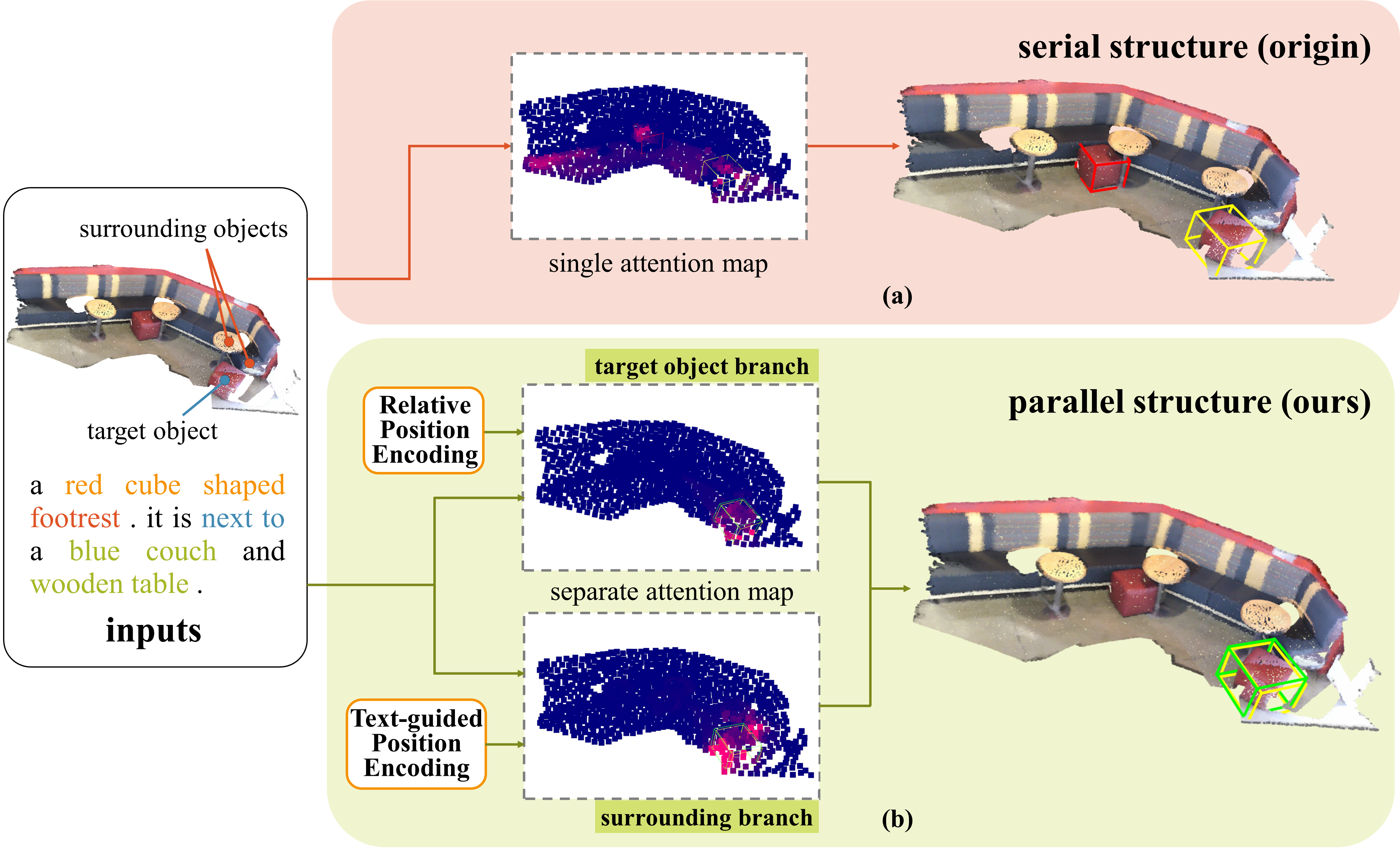} 
\caption{Comparison between the serial decoder structure (a) and our parallel decoder structure (b). For grounding results shown in scenes, yellow represents the ground-truth boxes, red is the wrongly localized box, and green is the successfully localized one. 
}
\label{fig:decoder}
\vspace{-8pt}
\end{figure}

However, existing methods have two problems.
\textbf{\textit{First}}, object attribute (e.g., shapes and colors) features and spatial layout features are entangled.
The former requires the network to pay more attention to points near the predicted boxes,
while the latter asks the attention map to focus more on points forming spatial relationships with target objects.
Previous serial structures cannot disentangle different types of attention.
For instance, as shown in Fig. \ref{fig:decoder} (a), the coupled features lead to a dispersed distribution of attention in the attention map, causing the model to locate objects of the same category but in different positions.
\textbf{\textit{Second}}, 
the spatial information from text tokens is not employed in the visual cross-attention block. 
Without the guidance of text information, queries can only roughly learn the features of neighboring points.
This causes queries to focus on redundant and irrelevant layout information, leading to a dispersion of attention.

To address these problems, we introduce a new framework named PD-APE.
A dual-branch decoder is employed to perform target object feature decoding and surrounding layout awareness in parallel.
Specifically, the target object branch refines the attention map through local visual features and text features that describe the target's attributes (e.g. category and color);
the surrounding branch captures the relationship between the target and surrounding objects through contextual text tokens and corresponding visual features.
Taking the text information into account, we semantically divide the tokens into target descriptions and layout descriptions, serving the target object branch and the surrounding branch, respectively.
Moreover, we design an adaptive position encoding method to enhance associated visual features.
In the target object branch, 
we obtain position encoding for each seed point by using its relative position to the surfaces of the predicted boxes.
It provides explicit information to guide the model to focus on the target object itself.
In the surrounding branch, we formulate the confidence between visual and text features as a gate to dominate the encoding process.
This gate guides the queries to focus on points that are related to the spatial layout described by the textual inputs.
The attention maps generated by our new framework are shown in Fig. \ref{fig:decoder} (b).

Our main contributions can be summarized as follows:
(a) We design a parallel decoder with two branches, each holding different attention maps to focus separately on target objects and surrounding layouts;
(b) We propose an adaptive position encoding method for the two branches of the decoder, which fully utilizes multi-modal inputs to guide the module to focus on points that carry effective information;
(c) Extensive experiments show that our PD-APE delivers new state-of-the-art performance on both ScanRefer and Nr3D benchmarks.

\section{Related Works}
\subsection{3D Vision-Language Tasks}

Vision and language, the two most essential and foundational modalities for machines to understand and interact with the 3D real world, give rise to various 3D vision-language tasks.
3D Dense Captioning (3DDC) \cite{chen2020scan2cap,yuan2022xtrans,chen2023endtoend} requires a model to accurately localize all objects in a complex 3D scene and generate descriptive captions for them. 
3D Visual Grounding (3DVG) \cite{chen2020scanrefer,achlioptas2020referit3d,jain2022bottom} takes 3D point clouds and language descriptions as inputs and generates corresponding bounding boxes to localize target objects.
3D Question Answering (3DQA) \cite{azuma2022scanqa,ma2023sqa3d} provides visual information from the 3D scenes for models, aiming to answer textual questions about the 3D scene.
All of the above tasks primarily focus on aligning visual and language features, encompassing object attributes (e.g., shapes and colors) and the relationships between targets and their environment.
In this work, we focus on 3D Visual Grounding tasks (3DVG), enabling machines to simultaneously comprehend inputs from 3D point clouds and natural language.

\subsection{3D Visual Grounding}

Previous works can be categorized into two-stage and one-stage methods based on their overall model structure.

The two-stage methods \cite{chen2020scanrefer,zhao20213dvg,yang2021sat,huang2022multiviewtransformer,guo2023viewrefer,luo20223d} first separately parse the language inputs and perform 3D point cloud object detection. 
In the second stage, the features of the extracted visual proposals and language tokens are fused, facilitating the identification of the best-matching objects.
Graph-based methods are widely used in early works \cite{achlioptas2020referit3d,huang2021text} to represent and process the spatial relationships between the detected proposals, achieving localization in 3D scenes.
Recent works \cite{zhao20213dvg,wang2023drpnet,yang2023exploiting,zhang2023multi3drefer,shi2024aware} employ transformers \cite{vaswani2017attention} as the key module, leveraging the cross-attention mechanism to achieve modality alignment and feature abstraction.

The one-stage methods \cite{jain2022bottom,wu2023eda,luo20223d} are proposed because, in the two-stage method, objects mistakenly overlooked during the pre-training stage cannot be learned or corrected during the fusion stage.
\cite{jain2022bottom} grounds referential utterances in point clouds by both top-down language guidance and bottom-up object guidance.
\cite{luo20223d} incrementally selects key points under language guidance and directly locates target objects with a goal-oriented mining module.
\cite{wu2023eda} explicitly decouples textual attributes in sentences and achieves dense alignment between language and point clouds through positional and semantic features.
These methods have achieved impressive results, but simultaneously learning the features of target objects and their surrounding layout can lead to a dispersed distribution of attention in the attention map.
To this end, we design a one-stage model with a parallel decoder, employing the position encoding method to guide the generation of the attention map and achieve precise grounding.

\begin{figure*}[!t]
\centering
\includegraphics[width=1.0 \textwidth]{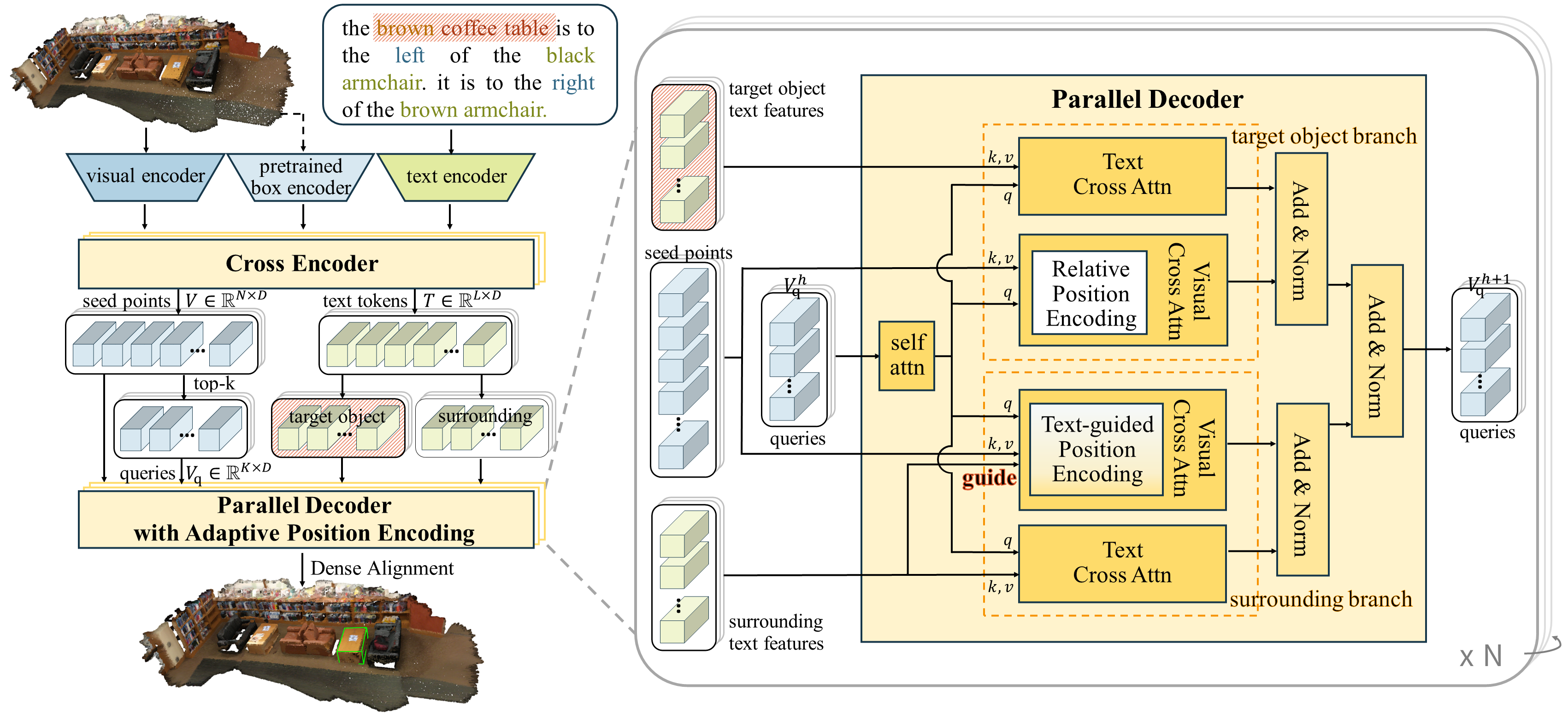} 
\caption{Architecture of our 3D Visual Grounding Framework.
With the inputs of natural languages and 3D point clouds, our model consists of three encoders for different modalities, a cross encoder, and a novel parallel decoder. 
Adaptive Position Encoding methods are added to cross-attention modules for both the target object branch and the surrounding branch. 
The final visual and text output features are aligned with each other to generate the detected boxes.
}
\label{fig:Framework}
\end{figure*}

\section{Methodology}
\subsection{Overview}

The objective of 3D visual grounding is to accurately localize objects in 3D point cloud scenes according to the given natural language descriptions. 
In this task, our model uses point cloud sampling to $N_{in}$ points as input, 
which includes
3 dimensions for localization (XYZ) and 3 dimensions for color (RGB). 
The input of texts is denoted as $T_{in}$ with length $L$.
The output is a set of boxes $B_{out} \in \mathbb{R}^{K_{out} \times 6}$, 
which represents an estimation of $K_{out}$ corresponding objects with 3 dimensions of box centers (XYZ) and 3 dimensions of box sizes (length-width-height).

The architecture of our framework is shown in Fig. \ref{fig:Framework}. 
Visual and text inputs first go through their respective modality encoders, then together pass through a cross encoder and a parallel decoder for target object features and surrounding features, and finally complete grounding through dense alignment following \cite{wu2023eda}.
Each module is illustrated in detail in this section.

\subsubsection{Modality Encoders.}  
The input texts are encoded by the pre-trained RoBERTa \cite{liu2019roberta}, which produces text tokens with features $T_{0} \in \mathbb{R}^{L \times D}$ and $D$ is the dimension of features.
For point cloud inputs, we provide two different backbone architectures: (a) PointNet++ \cite{qi2017pointnet++}; (b) a sparse 3D modification of ResNet34 \cite{he2015deepresidual} following FCAF3D \cite{rukhovich2022fcaf3d}. The former is adopted for a fair comparison with the recent methods, while the latter delivers better performance. The visual encoder samples $N$ seed points with features $V_{0} \in \mathbb{R}^{N \times D}$ from the point clouds.
Additionally, the GroupFree \cite{liu2021group} detector is optionally used to detect 3D boxes according to \cite{wu2023eda}, which are subsequently encoded as box tokens $B \in \mathbb{R}^{K \times D}$.

\subsubsection{Visual-Text Cross Encoder.}
We employ BUTD-DETR’s \cite{jain2022bottom} cross-encoder module for the intermodulation of visual and text features. 
Two branches with cross-attention blocks separately take features of one modality as query and features of another modality as key and value. 
The outputs of this module are represented as text token features $T \in \mathbb{R}^{L \times D}$ and visual features for seed points (seed point features) $V \in \mathbb{R}^{N \times D}$.
Then we select top-K candidates from seed points as queries through their predicted confidence scores, whose features are defined as $V_{q} \in \mathbb{R}^{K \times D}$.

\subsubsection{Parallel Decoder with Target Object Branch and Surrounding Branch.}
We design a decoder with two parallel branches. One is used to extract the target object features of query proposals, and another is used to perceive the environmental layout around each of them.
The query proposal features $V_{q}$ firstly pass a self-attention layer and act as the query of each cross-attention layer in each branch.

For the text features $T$, we divide the text tokens into five semantic components following \cite{wu2023eda}, using the off-the-shelf tool from \cite{schuster2015generating, wu2019unified}. 
Based on the five above components, target object tokens with features $T_{m}$ serve for the target object branch, and other tokens with features $T_{s}$ serve for the surrounding branch.
For instance, in \textit{``there is a dark brown wooden chair. placed in the table of the kitchen ."}, the \texttt{Main object} - \textit{``chair"} and the \texttt{Attributes} - \textit{``dark brown wooden"} are target object features $T_{m}$ and other parts \texttt{Auxiliary object}, \texttt{Pronoun}, and \texttt{Relationship} are surrounding features $T_{s}$. 
They separately act as the key and value of the text cross-attention layer in two branches.

The seed point features $V$ act as the key and value of the visual cross-attention layer in both two branches. 
Note that we design different adaptive position encoding methods for each branch, which will be described in detail in Sec. Adaptive Position Encoding.
Finally, visual-guided and text-guided features in each branch are fused together first, and then features from the two branches are fused. 
The query proposal features are updated as $V^{h}_{q} \in \mathbb{R}^{K \times D}$, where $h$ represents the $h-th$ layer of the decoder.

\begin{figure}[!t]
\centering
\includegraphics[width=0.49 \textwidth]{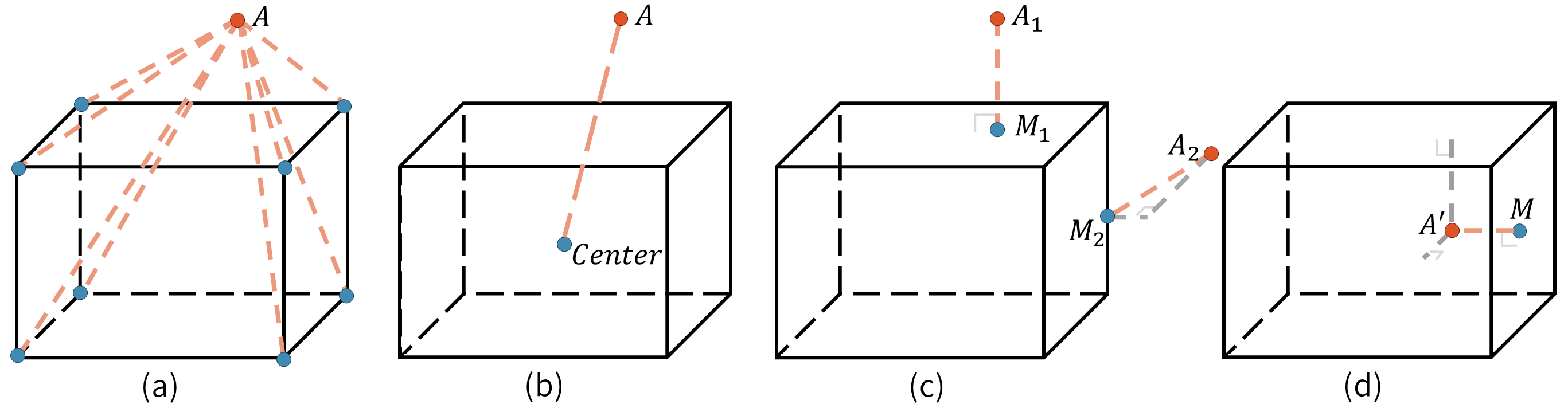} 
\caption{Different methods to calculate the relevant position between the sample of seed points A and the predicted box: (a) Vertex Relative Positioning \cite{shen2024vdetr}; (b) Center Relative Positioning; (c)(d) Box-surface Relative Positioning for points outside/inside the box.
}
\label{fig:Position_Encoding}
\vspace{-8pt}
\end{figure}

\subsubsection{Prediction Head for Features and Boxes.}
Following \cite{wu2023eda}, the query proposal features $V^{h}_{q}$ in each layer are projected into position features $V^{h}_{l} \in \mathbb{R}^{K \times D_{l}}$ and object semantic features $V^{h}_{o} \in \mathbb{R}^{K \times D_{o}}$ for losses calculation, where $D_{l}$ and $D_{o}$ represents the dimensions of two types of features. 
A box prediction head \cite{liu2021group} is employed to generate bounding boxes $B_{out} \in \mathbb{R}^{K_{out} \times 6}$.

\subsection{Adaptive Position Encoding}
\label{sec:pe}
\subsubsection{Position Encoding for Attention Map.}
Position encoding (PE) can effectively refine the attention map, which is crucial for understanding the contextual information of points in 3D scenes.
For the 3D visual grounding task, we update the calculation method of the attention map for the decoder as follows:
\begin{equation}
\label{eq:modulate_cross_attn}
  \widehat{\mathcal{A}} = \operatorname{softmax}(\mathcal{Q}\mathcal{K}^{\text{T}}{\;+ \;\mathcal{E}}),
\end{equation}
where $\mathcal{Q}$ and $\mathcal{K}$ represent the query features and the key-value features respectively and $\mathcal{E}$ represents the position encoding results.

We encode the relative positions between seed points and coarsely predicted boxes as follows:
\begin{equation}
\label{eq.rpe_mlp_4}
{\mathcal{E}} = \operatorname{MLP}(\operatorname{F}(\Delta\mathcal{E}),
\end{equation}
where $\operatorname{F}(\cdot)$ is a non-linear function and $\Delta\mathcal{E} \in \mathbb{R}^{K \times N \times 3}$ is the calculated relative position. 
$\operatorname{MLP}$ represents an MLP-based transformation that projects the
features to a higher dimension space.
For a detailed explanation of this part, please refer to \cite{shen2024vdetr}. 

\subsubsection{Box-surface Relative Positioning Method.}
To represent the relative position between $N$ seed points and $K$ boxes, we find the point on the box surface closest to each seed point and use the offsets in three dimensions as $\Delta\mathcal{E} = \{(e^{ij}_{x}, e^{ij}_{y}, e^{ij}_{z}) |\quad i = 0,1,...,K ; j = 0,1,...,N\}$, which is shown in Fig. \ref{fig:Position_Encoding} (c)(d).
To be specific, for one of seed points $A = (x_{0}, y_{0}, z_{0})$, our goal is to find a point $M = (x, y, z)$ on the box surface to minimize the distance between $M$ and $A$. The offset can be set as
\begin{equation}
\label{eq.rel_dist_p}
\begin{split}
    (e_{x}, e_{y}, e_{z}) = 
        (x_{0} - x, y_{0} - y, z_{0} - z).
\end{split}
\end{equation}
We set no absolute values here in order to distinguish between points inside and outside the box with the same offset.

For a point $A$ outside the box as Fig. \ref{fig:Position_Encoding} (c), with a box $b = (x_{c}, y_{c}, z_{c}, l, w, h)$ 
the three-dimensional offset can be represented as
\begin{equation}
\label{eq.rel_dist_all}
\begin{split}
    (e_{x}, e_{y}, e_{z}) = 
        (\operatorname{max}(|x_{0}-x_{c}|-\frac{l}{2}, 0), \\
        \operatorname{max}(|y_{0}-y_{c}|-\frac{w}{2}, 0), 
        \operatorname{max}(|z_{0}-z_{c}|-\frac{h}{2}, 0)).
\end{split}
\end{equation}

When the seed point $A' = (x'_{0}, y'_{0}, z'_{0})$ is inside the box as Fig. \ref{fig:Position_Encoding} (d), the minimum distance of $M$ and $A'$ can be represented as:
\begin{equation}
    \label{eq.rel_dist_min_}
    MA'_{min} = \operatorname{min}(\frac{l}{2}-|x'_{0}-x_{c}|, 
    \frac{w}{2}-|y'_{0}-y_{c}|, \frac{h}{2}-|z'_{0}-z_{c}|).
\end{equation}
If $MA'$ reaches its minimum value in the x-direction, then $e_{x}$ is set to $-MA'$, and the other two offsets are set to 0 because $M$ and $A'$ have the same coordinate values in the y-direction and z-direction; the same applies to $e_{y}$ and $e_{z}$.
The offset takes the negative of the distance to differentiate from the seed point outside the box.
For detailed mathematical reasoning, please refer to the supplementary material.

Compared to the method shown in Fig. \ref{fig:Position_Encoding} (b) that directly uses the offset between the point and the box center as the relative position, 
our Box-surface Relative Positioning method incorporates both relative distance information and the shape information of the box. 
In contrast to the Vertex Relative Positioning method proposed by \cite{shen2024vdetr}, as shown in Fig. \ref{fig:Position_Encoding} (a), which uses eight MLP-based transformations in each layer, our method encodes only once per layer of cross-attention, making it more time-efficient and memory-saving.
For detailed ablation experiments, please refer to Sec. Experiments.

\begin{figure}[!t]
\centering
\includegraphics[width=0.45 \textwidth]{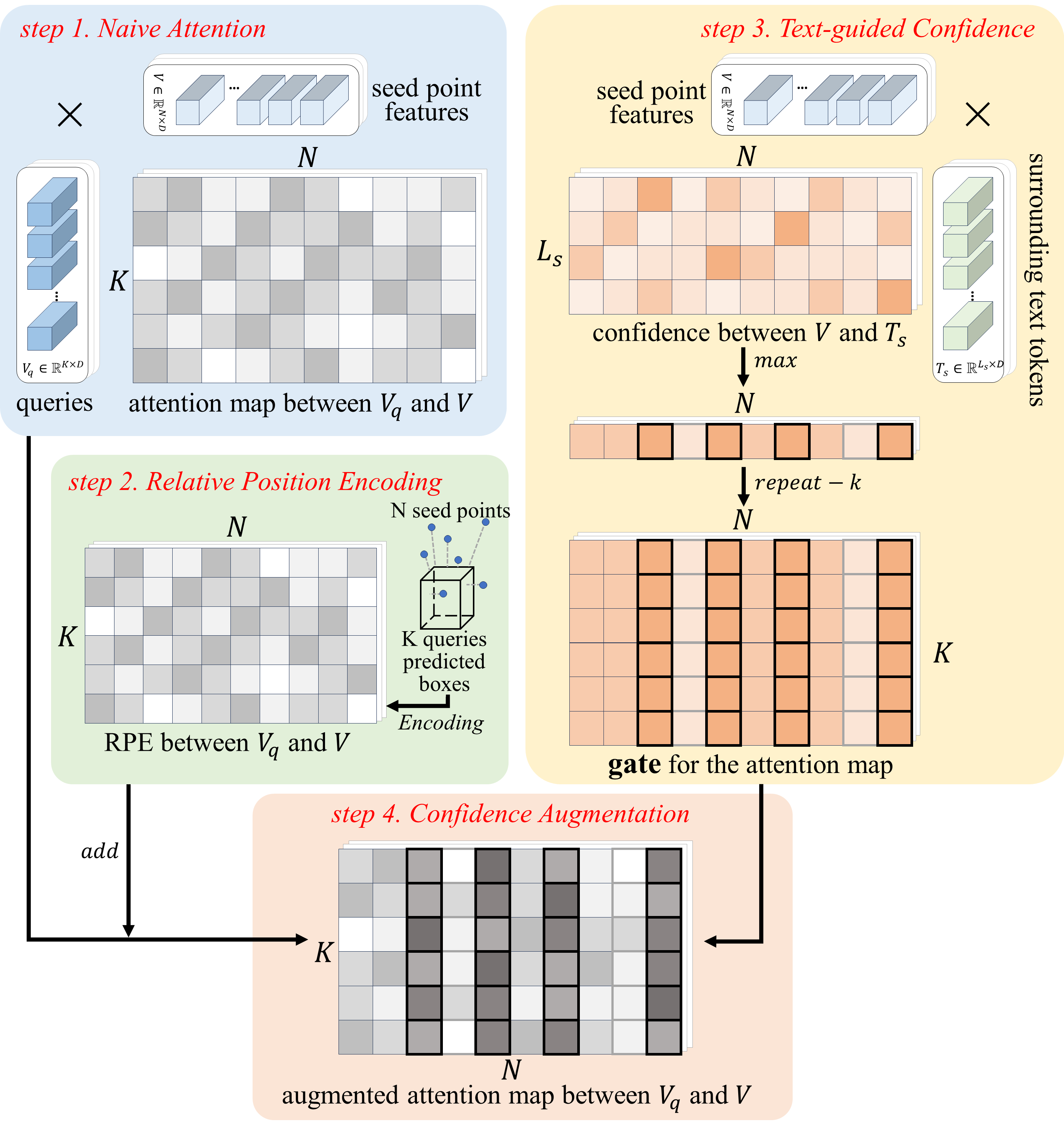} 
\caption{The illustration of the Adaptive Position Encoding method. Step 3 is the text-guided confidence that is only used in the surrounding branch.}
\label{fig:tpe}
\vspace{-8pt}
\end{figure}

\begin{table*}[!t]
    \small
    \centering
    \setlength{\tabcolsep}{5pt}
    \renewcommand{\arraystretch}{1.0}
    \setlength{\abovecaptionskip}{2pt}
    \begin{tabular}{lc|cc|cc|cc}
    \toprule
        &    & \multicolumn{2}{c|}{Unique ($\sim$19\%)} & \multicolumn{2}{c|}{Multiple ($\sim$81\%)} & \multicolumn{2}{c}{\textbf{Overall}} \\
    \multirow{-2}{*}{Method}                   & \multirow{-2}{*}{Venue} & @0.25  & @0.5   & @0.25  & @0.5   & @0.25 & @0.5     \\ 
    \midrule
    ScanRefer~\cite{chen2020scanrefer}         & ECCV2020                & 67.6 & 46.2 & 32.1 & 21.3 & 39.0 & 26.1 \\
    ReferIt3D~\cite{achlioptas2020referit3d}   & ECCV2020                & 53.8 & 37.5 & 21.0 & 12.8 & 26.4 & 16.9  \\ 
    TGNN~\cite{huang2021text}                  & AAAI2021                & 68.6 & 56.8 & 29.8 & 23.2 & 37.4 & 29.7 \\ 
    InstanceRefer~\cite{yuan2021instancerefer} & ICCV2021                & 77.5 & 66.8 & 31.3 & 24.8 & 40.2 & 32.9 \\ 
    SAT~\cite{yang2021sat}   \dag              & ICCV2021                & 73.2 & 50.8 & 37.6 & 25.2 & 44.5 & 30.1 \\ 
    FFL-3DOG~\cite{feng2021ffl}                & ICCV2021                & 78.8 & 67.9 & 35.2 & 25.7 & 41.3 & 34.0 \\ 
    3DVG-Transformer~\cite{zhao20213dvg}       & ICCV2021                & 77.2 & 58.5 & 38.4 & 28.7 & 45.9 & 34.5 \\
    3D-SPS~\cite{luo20223d}    \dag            & CVPR2022                & 84.1 & 66.7 & 40.3 & 29.8 & 48.8 & 37.0 \\ 
    3DJCG~\cite{cai20223djcg}                  & CVPR2022                & 78.8 & 61.3 & 40.1 & 30.1 & 47.6 & 36.1 \\
    BUTD-DETR~\cite{jain2022bottom}            & ECCV2022                & 82.9 & 65.0 & 44.7 & 34.0 & 50.4 & 38.6 \\ 
    D3Net~\cite{chen2022d3net} \dag            & ECCV2022                & -    & 70.4 & -    & 30.5 & -    & 37.9 \\ 
    EDA~\cite{wu2023eda}                       & CVPR2023                & 85.8 & 68.6 & 49.1 & 37.6 & 54.6 & 42.3 \\ 
    ViewRefer~\cite{guo2023viewrefer}          & ICCV2023                & -    & -    & 33.1 & 26.5 & 41.3 & 33.7 \\ 
    3DRP-Net~\cite{wang2023drpnet}             & EMNLP2023               & 83.1 & 67.7 & 42.1 & 32.0 & 50.1 & 38.9 \\ 
    CORE-3DVG~\cite{yang2023exploiting}        & NIPS2023                & 85.0 & 67.1 & 51.8 & 39.8 & 56.8 & 43.8 \\ 
    3DVLP~\cite{zhang20233dvlp}                & AAAI2024                & 85.2 & 70.0 & 43.7 & 33.4 & 51.7 & 40.5 \\
    VPP-Net~\cite{shi2024aware}                & CVPR2024                & 86.1 & 67.1 & 50.3 & 39.0 & 55.7 & 43.3 \\
    \midrule
    \textbf{PD-APE (ours-PointNet++)}          &                         & 86.7 & 73.2 & 50.2 & 40.6 & 56.6 & 45.5 \\
    \textbf{PD-APE (ours-Sparse3D)}            &                         & \textbf{87.7} & \textbf{75.1} & \textbf{52.6} & \textbf{42.2} & \textbf{57.8} & \textbf{47.1} \\
    \bottomrule
    \end{tabular}
    \caption{The 3D visual grounding results on ScanRefer, accuracy evaluated by IoU 0.25 and IoU 0.5.
    Methods with \dag means that they train and evaluate with 2D information assistance. 
    All the above results shown in the table are the optimal accuracy of each method as reported in its paper.}
    \label{tab:ScanRefer_result}
    \vspace*{-1\baselineskip}
\end{table*}

\subsubsection{Text-guided PE for Surrounding Branch.}

The PE method mentioned above uses relative positions alone to provide confidence for the attention map.
This is inaccurate in the surrounding branch, because it cannot determine whether the seed point carries valid surrounding layout information.
As shown in Fig. \ref{fig:tpe}, we propose a text-guided method as gates to refine the attention map and update Eq. \ref{eq:modulate_cross_attn} as follows:
\begin{equation}
\label{eq:modulate_cross_attn_b}
  \widehat{\mathcal{A}} = \operatorname{softmax}({gate}(\mathcal{Q}\mathcal{K}^{\text{T}}{\;+ \;\mathcal{E}})).
\end{equation}

To be specific, we first compute the confidence between seed point features $V \in \mathbb{R}^{N \times D}$ and surrounding text tokens' features $T_{s} \in \mathbb{R}^{L_{s} \times D}$:
\begin{equation}
\label{eq:confidence}
  confidence = V(T_{s})^{\text{T}},
\end{equation}
where $confidence \in \mathbb{R}^{N \times {L_{s}}}$. 
Then we take the maximum confidence for each seed point with one of $L_{s}$ text tokens:
\begin{equation}
\label{eq:gate}
  gate_{0} = \sigma(\operatorname{max}(confidence)),
\end{equation}
where $gate_{0} \in \mathbb{R}^{N}$. 
$\sigma$ is the sigmoid function, which is employed to map the maximum confidence to a range between 0 and 1.
With higher confidence, the gate value approaches 1, allowing the corresponding value in the attention map to be effectively updated by the seed point; conversely, it approaches 0 and the update of the attention map is ignored.
Finally, we repeat $gate_{0}$ for $K$ times to generate 
$gate \in \mathbb{R}^{N \times K}$ and augment the attention map between seed point features and query features.

\subsection{Training Objectives}
Following \cite{wu2023eda}, we use position and semantic loss for dense alignment, as follows:
\begin{equation}
\label{eq:loss}
  \mathcal{L}_{all} = \mathcal{L}_{pos} + \mathcal{L}_{sem}.
\end{equation}
Please refer to \cite{wu2023eda} for more details.

\section{Experiments}
\subsection{Experiment settings}

We perform experiments on two commonly used datasets, \textbf{ScanRefer} \cite{chen2020scanrefer} and \textbf{Nr3D} \cite{achlioptas2020referit3d}. 
Both datasets are derived from ScanNet \cite{dai2017scannet}, which contains 1201 indoor 3D scenes for training and 312 for validation.
ScanRefer provides 51,583 human-annotated language descriptions of 11,046 objects in 800 3D scenes, which is officially divided into a 36,665-sample training set and a 9,508-sample testing set.
Nr3D contains 41,503 language annotations, describing 76 object categories from 707 indoor scenes.

For ScanRefer, we adopt Acc@mIoU as the evaluation metric, where m is set as 0.25 and 0.5. 
This metric indicates the proportion of predicted bounding boxes that have an Intersection over Union (IoU) greater than m with the ground-truth bounding boxes. 
For Nr3D, experiments are all evaluated under the Acc@0.25IoU.

\begin{table*}[!t]
    \small
    \centering
    \setlength{\tabcolsep}{5pt}
    \renewcommand{\arraystretch}{1.0}
    \setlength{\abovecaptionskip}{4pt}
    \begin{tabular}{l|cc|cc|c|cc|cc|c}
        \toprule
                     & \multicolumn{5}{c|}{Ground Truth Protocol} & \multicolumn{5}{c}{Classifier Protocol}         \\
        \multirow{-2}{*}{Method}      & Easy & Hard & V-Dep & V-Indep & \textbf{Overall} & Easy & Hard & V-Dep & V-Indep & \textbf{Overall} \\
        \midrule
       ReferIt3D~\cite{achlioptas2020referit3d}   & 43.6 & 27.9 & 32.5 & 37.1 & 35.6 & -    & -    & -    & -    & -    \\
       TGNN~\cite{huang2021text}                  & 44.2 & 30.6 & 35.8 & 38.0 & 37.3 & -    & -    & -    & -    & -    \\
       InstanceRefer~\cite{yuan2021instancerefer} & 46.0 & 31.8 & 34.5 & 41.9 & 38.8 & -    & -    & -    & -    & -    \\
       3DVG-Trans~\cite{zhao20213dvg}       & 48.5 & 34.8 & 34.8 & 43.7 & 40.8 & -    & -    & -    & -    & -    \\ 
       SAT~\cite{yang2021sat}                     & 56.3 & 42.4 & 46.9 & 50.4 & 49.2 & -    & -    & -    & -    & -    \\
       MVT~\cite{huang2022multiviewtransformer}   & 61.3 & 49.1 & 54.3 & 55.4 & 55.1 & -    & -    & -    & -    & -    \\ 
       BUTD-DETR~\cite{jain2022bottom}            & -    & -    & -    & -    & -    & 60.7 & 48.4 & 46.0 & 58.0 & 54.6 \\
       ViL3DRel~\cite{wu2023eda}                  & 70.2 & 57.4 & 62.0 & 64.5 & 64.4 & -    & -    & -    & -    & -    \\
       EDA~\cite{wu2023eda}                       &      &      &      &      &      & 58.2 & 46.1 & 50.2 & 53.1 & 52.1 \\
       ViewRefer~\cite{guo2023viewrefer}          & 63.0 & 49.7 & 55.1 & 56.8 & 56.0 & -    & -    & -    & -    & -    \\
       3DRP-Net~\cite{wang2023drpnet}             & 71.4 & \textbf{59.7} & 64.2 & 65.2 & 65.9 & -    & -    & -    & -    & -   \\ 
       VPP-Net~\cite{shi2024aware}                & -    & -    & -    & -    & -    & -    & -    & -    & -    & 56.9 \\
       \midrule
       \textbf{PD-APE(Ours)}                      & \textbf{77.0} & 58.5 & \textbf{64.4} & \textbf{70.1} & \textbf{67.8} & \textbf{67.7} & \textbf{50.6} & \textbf{52.8} & \textbf{62.3} & \textbf{59.1} \\ 
       \bottomrule
    \end{tabular}
    \caption{Performance on Nr3D datasets by Acc@0.25IoU as the metric. 
    There are some previous methods that only provide evaluation results on the dataset with detected lables, 
    which are not shown in this table.}
    \label{tab:NR3D_result}
    \vspace*{-1\baselineskip}
\end{table*}

\begin{table}[!t]
    \small
    \centering
    \setlength{\abovecaptionskip}{2pt}
    \begin{tabular}{c|c|c|c|c}
    \toprule
     \multicolumn{1}{l|}{}& PE Method &  @0.5 & Latency & GPU Memory      \\ 
    \midrule
     (a) & -           & 42.3     & 118 ms    & 3416 MB \\
     (b) &Vertex RPE           & 44.5     & 155 ms   & 4168 MB \\
     (c) &Center RPE              & 43.0    & 121 ms  & 3668 MB   \\
     (d) &Box-surface RPE   & 44.5  & 123 ms & 3672 MB \\ 
    \bottomrule
    \end{tabular}
    \caption{Evaluation results and inference costs of different types of PE methods.
    For Latency costs per scene and the GPU memory costs, we evaluate on one NVIDIA 4090 24GB GPU with batch size as 1 for a fair comparison.}
    \label{tab:Ablation_pe}
    \end{table}

\subsection{Quantitative Comparisons}

Tab. \ref{tab:ScanRefer_result} exhibits our experiments on the ScanRefer dataset compared with previous works.
Our PD-APE with the backbone of PointNet++ outperforms all the previous methods on Acc@0.25IoU and Acc@0.5IoU, achieving 56.6\% and 45.5\%, respectively.
It surpasses our baseline EDA \cite{wu2023eda} by 3.2\% Acc@0.5IoU, and also 1.7\% higher than that of the previous state-of-the-art method \cite{yang2023exploiting}.
Replacing the backbone with Sparse3D ResNet34, our PD-APE reaches even higher results, 57.8\% Acc@0.25IoU and 47.1\% Acc@0.5IoU.
Note that compared to methods using 2D+3D inputs, our approach achieves even better results without relying on additional 2D features for assistance.

We also report our experiment results with PointNet++ on the Nr3d dataset. 
Different from ScanRefer with detected boxes as ground truth, the grounding task on Nr3d aims to locate target objects from all the given ground truth bounding boxes.
In the standard protocol from ReferIt3D \cite{achlioptas2020referit3d}, the ground truth labels for boxes are adopted to independently assess the model's grounding ability.
BUTD-DETR \cite{jain2022bottom} uses ground truth boxes with labels from the pointnet++ classifier. 
For a fair comparison, we provide the evaluation results under both protocols separately in Tab. \ref{tab:NR3D_result}.
During the training phase, we employ the data augmentation method in ViL3DRel \cite{chen2022vil3dref}
in the ground truth protocol, and use the data augmentation method following BUTD-DETR \cite{jain2022bottom} in the classifier protocol for fairness.

In the ground truth protocol, our PD-APE performs better than all the previous methods and reaches 67.8\%, surpassing the previous state-of-the-art method \cite{wang2023drpnet} by 1.9\%.
In the classifier protocol, PD-APE achieves an overall accuracy of 59.1\%, 2.2\% higher than the current best result reported by VPP-Net \cite{shi2024aware} and 7.0\% higher than our baseline EDA \cite{wu2023eda}.

\begin{table}[!t]
    \small
    \centering
    \renewcommand{\arraystretch}{1.0}
    \setlength{\abovecaptionskip}{2pt}
    \setlength{\tabcolsep}{8pt}
    \begin{tabular}{c|c|c|c|c}
    \toprule
    & Decoder        &  &  &      \\ 
    & Structure &\multirow{-2}{*}{@0.5} &\multirow{-2}{*}{Latency/scene}  &\multirow{-2}{*}{GPU Memory} \\
    \midrule
    (a) & Serial   & 42.3    & 118 ms    & 3416 MB     \\
    (b) & Parallel & 43.3    & 125 ms    & 3644 MB     \\
    \bottomrule
    \end{tabular}
    \caption{Evaluation results of different decoder structures.
    For Latency costs per scene and the GPU memory costs, we evaluate on one NVIDIA 4090 24GB GPU with batch size as 1 for a fair comparison.}
    \label{tab:Ablation_stru}
    \vspace{-10pt}
    \end{table}
    
\subsection{Ablation Studies}
\label{sec:as}
We perform ablation studies to understand the effectiveness of each component of our PD-APE.
Without further specification, all experiments are conducted on ScanRefer with 400 epochs, employing PointNet++ as the backbone.

\subsubsection{Comparison of Different Relative Positioning Methods.}
In Tab. \ref{tab:Ablation_pe}, we first compare the effect of different Relative Positioning methods.
The experiment (a) is our baseline EDA \cite{wu2023eda}.
Keeping the structure of the model, we add PE with different types of relative positioning methods to the decoder's cross-attention block.
The experiment (b) use Vertex Relative PE following \cite{shen2024vdetr}, as shown in Fig. \ref{fig:Position_Encoding} (a), achieving 44.5\% (+2.2\%) Acc@0.5IoU.
As shown in experiment (c), the Center Relative PE method (Fig. \ref{fig:Position_Encoding} (b)) results in limited improvement to Acc@0.5IoU (43.0\%, +0.7\%), because the method only considers the relative positional relationship between seed points and query proposals, without taking the shape and size of the predicted boxes into account.
We finally use the Box-surface PE method in experiment (d) (Fig. \ref{fig:Position_Encoding} (c) \&(d)), which reaches comparable results as experiment (b).
Considering the large latency and unavoidable memory costs brought by Vertex RPE, we take the Box-surface RPE as our priority.

\subsubsection{Comparison of Different Structures of Decoders.}
We provide the ablation study on the structure of the decoder and illustrate the results in Tab. \ref{tab:Ablation_stru}.
Comparing the experiment (a) and (b), our design that changes the decoder's serial structure to a dual-branch parallel structure increases the Acc@0.5IoU from 42.3\% to 43.3\% (+1.0\%).
This proves that allowing the decoder to separately focus on the features of the query proposals and their surrounding layouts can lead grounding more accurate.

\begin{figure*}[!t]
\centering
\includegraphics[width=1.0 \textwidth]{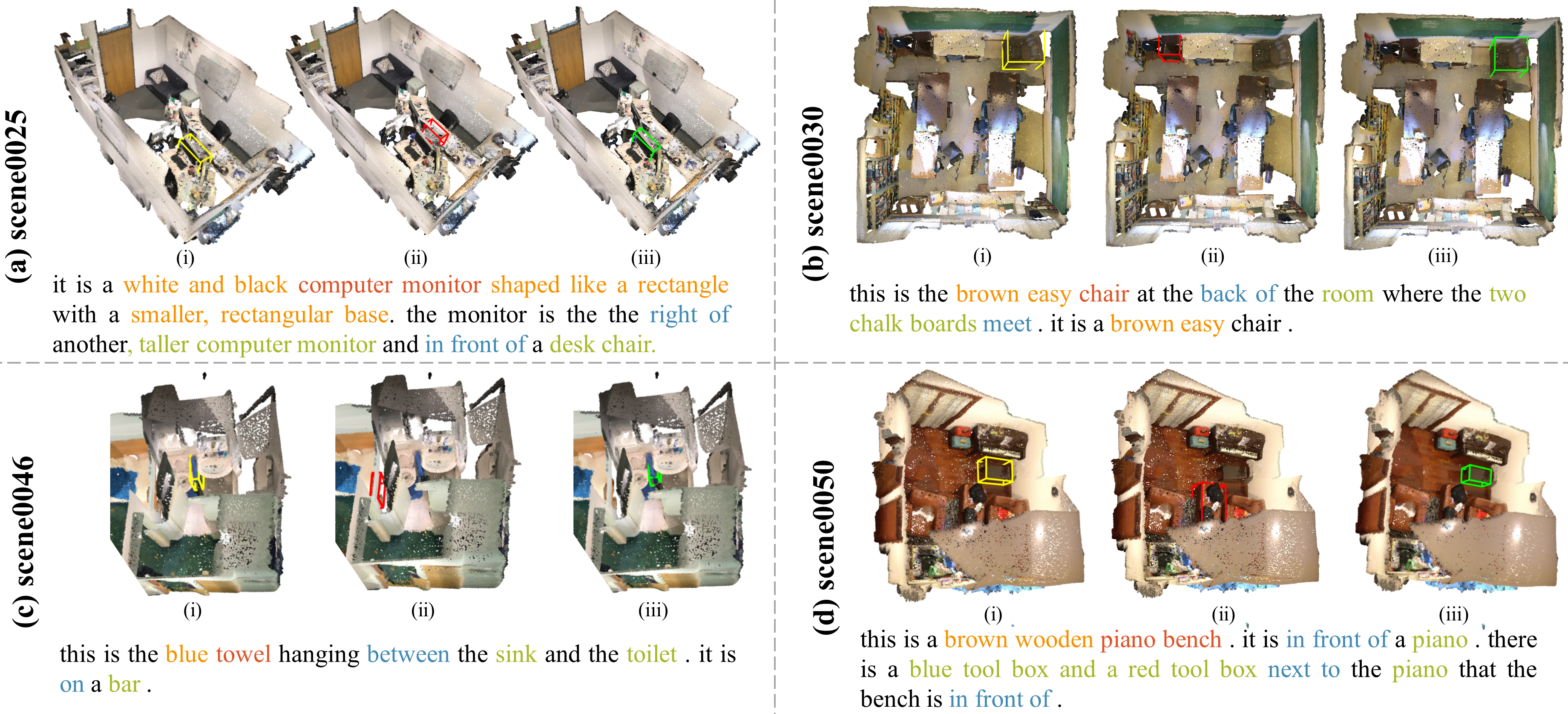} 
\caption{Visualization results of different models for scenes from the ScanRefer dataset. For all boxes, yellow represents the ground-truth references; red represents results from EDA that contain grounding errors; green represents proposals generated by our PD-APE.
Words in different colors show the results of text decoupling.
}
\label{fig:visual}
\vspace{-4pt}
\end{figure*}

\subsubsection{Effect of Text-guided PE Methods on Two Branches.}
To verify the effectiveness of the PE for our dual-branch decoder, 
we conduct ablation experiments and present the results in Tab. \ref{tab:Ablation_tg}.
In experiment (a), we adopt the parallel decoder with the naive attention method and no position encoding, which reaches 43.3\% Acc@0.5IoU.
After equipping Box-surface RPE in both branches in experiment (b), the evaluation result of Acc@0.5IoU achieves 45.0\% (+1.7\%).
Then in experiment (c), we add a text-guided gate to the PE for the surrounding branch in the decoder, 
which increases the evaluation result to 45.5\% Acc@0.5IoU.
Note that we also add the text-guided gate in both branches, but the Acc@0.5IoU only reaches 44.8\% in experiment (d).
The decrease occurs because, for the target object branch, the decoder overlooks points that are incorrectly filtered out by the gate, thus failing to fully learn the query features.

\begin{table}[!t]
    \small
    \centering
    \renewcommand{\arraystretch}{1.0}
    \setlength{\abovecaptionskip}{2pt}
    \setlength{\tabcolsep}{8pt}
    \begin{tabular}{c|cc|cc}
    \toprule
    \multicolumn{1}{l|}{}& PE for t-branch & PE for s-branch      & @0.25 & @0.5      \\ 
    \midrule
    (a) &  & & 54.4          & 43.3          \\
    (b) & \checkmark & \checkmark & 55.2 & 45.0          \\
    (c) & \checkmark & \checkmark \S & \textbf{56.6} & \textbf{45.5}  \\
    (d) & \checkmark \S & \checkmark \S & 55.8 & 44.8          \\
    \bottomrule
    \end{tabular}
    \caption{Evaluation results of Adaptive Position Encoding methods for different branches.
    \checkmark means naive PE method and \S means Text-guided PE method.}
    \label{tab:Ablation_tg}
    \vspace{-10pt}
    \end{table}

\subsubsection{Effect of Different Text-guided Augmentations.}
In this section, we set experiments to evaluate how text-guided confidence can best augment the attention map. 
In Tab. \ref{tab:Ablation_gate}, $attn$ represents the attention map, $PE$ represents the relative position encoding, and $conf$ is the confidence between text tokens and seed points, which is described in detail in Sec. Methodology. 
Based on experiment (a) with no text-guided confidence, we try to add confidence as bias in experiment (b), and as the gate for PE and the entire attention map in experiment (c) and (d), separately.
The results indicate that using confidence as a gate on the entire attention map can best guide queries to attend to seed points that provide effective layout information, achieving 45.5\% Acc@0.5IoU.

\begin{table}[!t]
    \small
    \centering
    \renewcommand{\arraystretch}{1.0}
    \setlength{\abovecaptionskip}{2pt}
    \setlength{\tabcolsep}{8pt}
    \begin{tabular}{c|c|cc}
    \toprule
    \multicolumn{1}{l|}{}& Text-guided method      & @0.25 & @0.5      \\ 
    \midrule
    (a) & - & 55.2 & 45.0          \\
    (b) & $attn + PE + conf $ & 55.8 & 45.1          \\
    (c) & $attn + conf \times PE$  & 55.6 & 45.2          \\
    (d) & $conf \times (attn + PE)$ & \textbf{56.6} & \textbf{45.5}  \\
    \bottomrule
    \end{tabular}
    \caption{Evaluation results of Different Text-guided method.
    Experiments above all employ the parallel decoder and only use the Text-guided method in the surrounding branch.
    }
    \label{tab:Ablation_gate}
    \vspace{-10pt}
    \end{table}

\subsection{Visualization}
In Fig. \ref{fig:visual}, we exhibit the visualization results of 4 scenes from the ScanRefer dataset.
The boxes in colors of yellow, red, and green represent the ground-truth boxes, the top-1 predicted boxes generated by EDA \cite{wu2023eda}, and boxes generated by our PD-APE.
The results demonstrate the effectiveness of our method in understanding contextual information in the text to identify the target object.
On the one hand, it can avoid errors in extracting features such as colors and shapes (e.g., Fig. \ref{fig:visual} (a) identifies the black-and-white monitor as the white one), as well as errors in object category recognition(e.g., Fig. \ref{fig:visual} (d) recognizes ottoman as bench).
On the other hand, it can effectively understand the spatial layout around the target proposals, thereby correctly positioning the described object (e.g., Fig. \ref{fig:visual} (b)(c)).

\section{Conclusion}
In this paper, we presented PD-APE, a one-stage decoding framework for the 3D Visual Grounding task, which helps extract features of proposals and perceive the spatial layouts around them separately.
A novel Adaptive Position Encoding method was proposed to 
guide the generation of the attention map.
It is dominated by the relative position between seed points and predicted boxes in the target object branch, and facilitated by the text tokens with effective layout information in the surrounding branch.
Comprehensive experiments demonstrated the effectiveness of our method.

\bigskip

\bibliography{main}

\end{document}